\documentclass{llncs}
\usepackage{amssymb,amsfonts,amsmath}
\usepackage{times}
\usepackage{latexsym}
\usepackage{tikz,pgf}
\tikzstyle{arg}=[draw, thick, circle]
\usetikzlibrary{arrows}
\usepackage{subfigure}
\usepackage{a4,a4wide}
\newcommand{\AS}{\mathcal{AS}}
\newcommand{\body}[1]{B(#1)}
\newcommand{\bodyp}[1]{B^+(#1)}
\newcommand{\bodyn}[1]{B^-(#1)}
\newcommand{\BU}{{\ensuremath{B_{\cal U}}}}
\newcommand{\commadots}[0]{,\ldots ,}
\newcommand{\derives}{\la}
\newcommand{\GP}{\ensuremath{Gr(\p)}}
\newcommand{\head}[1]{H(#1)}
\newcommand{\la}{\leftarrow}
\newcommand{\naf}{{\it not}\,}
\newcommand{\p}{\ensuremath{{\pi}}}
\newcommand{\U}{{\ensuremath{\cal U}}}
\newcommand{\UP}{\ensuremath{U_{\p}}}
\newcommand{\sccs}{\mathit{SCCs(F)}}
\newcommand{\minimize}{\#\mathit{minimize}}

\newcommand{\inA}{\mathrm{in}}
\newcommand{\out}{\mathrm{out}}
\newcommand{\inS}{\mathrm{inS}}
\newcommand{\att}{\mathrm{defeat}}
\newcommand{\defeat}{\att}
\renewcommand{\succ}{\mathrm{succ}}
\renewcommand{\inf}{\mathrm{inf}}
\renewcommand{\sup}{\mathrm{sup}}

\newcommand{\order}{<}

\newcommand{\pgroundN}{\mathit{groundN}}

\newcommand{\argset}{\mathrm{arg\_set}}
\newcommand{\pargset}{\mathit{arg\_set}}
\newcommand{\inU}{\mathrm{inU}}
\newcommand{\defeatN}{\mathrm{defeatN}}
\newcommand{\defended}{\mathrm{defendedN}}

\newcommand{\cut}{\mathrm{notInSplusN}} %
\newcommand{\pcut}{\mathit{F\_minus\_range}}
\newcommand{\alphaN}{\mathrm{in\_SplusN}}
\newcommand{\uAlpha}{\mathrm{u\_cap\_Splus}}
\newcommand{\mr}{\mathrm{mr}}
\newcommand{\pmr}{\mathit{MR}}
\newcommand{\reach}{\mathrm{reach}}
\newcommand{\nmr}{\mathrm{self\_defeat}}
\newcommand{\nsym}{\mathrm{nsym}}
\newcommand{\reachnotvia}{\mathrm{reachnotvia}}
\newcommand{\cyc}{\mathrm{cyc}}
\newcommand{\bad}{\mathrm{bad}}
\newcommand{\posmr}{\mathrm{pos\_mr}}
\newcommand{\notminimal}{\mathrm{notminimal}}

\newcommand{\pstableN}{\mathit{stableN}}
\newcommand{\nemptyT}{\mathrm{nemptyT}}
\newcommand{\emptyT}{\mathrm{emptyT}}
\newcommand{\exitsMR}{\mathrm{existsMR}}

\newcommand{\true}{\mathrm{true}}
\newcommand{\defeated}{\mathrm{defeated}}
\newcommand{\notdefeated}{\mathrm{not\_in\_rangeN}}

\newcommand{\piterate}{\mathit{iterate}}
\newcommand{\twplus}{\mathrm{t\_mrOplus}}
\newcommand{\notmr}{\mathrm{not\_exists\_mr}}
\newcommand{\pispoilstage}{\mathit{satstage}}
\newcommand{\inN}{\mathrm{inN}}
\newcommand{\outN}{\mathrm{outN}}
\newcommand{\spoil}{\mathrm{fail}}
\newcommand{\eqplus}{\mathrm{eqplus}}
\newcommand{\equalplus}{\mathit{eq}^+}

\newcommand{\notdefended}{\mathrm{undefended}}
\newcommand{\eqplusupto}{\mathrm{eqp\_upto}}
\newcommand{\notdefeatedupto}{\mathrm{undefeated\_upto}}
\newcommand{\pdefended}{\mathit{defendedN}}
\newcommand{\I}{{\cal I}}
\renewcommand{\S}{{\cal S}}
\newcommand{\corr}{\cong}
\newcommand{\inSplus}{\mathrm{in\_range}}
\newcommand{\inRangeN}{\mathrm{in\_rangeN}}
\newcommand{\rangeN}{\mathit{rangeN}}

\newcommand{\Pol}{{\rm P}}

\newcommand{\NP}{\mbox{\rm NP}}

\newcommand{\co}{\mbox{\rm co}}
\newcommand{\CONP}{\mbox{\rm coNP}}
\newcommand{\coNP}{\mbox{\rm coNP}}

\newcommand{\SigmaP}[1]{{\rm \Sigma}_{#1}^{P}}
\newcommand{\PiP}[1]{{\rm \Pi}_{#1}^{P}}

\newcommand{\cf}{\mathit{cf}}
\newcommand{\stable}{{\mathit{stb}}}
\newcommand{\adm}{\mathit{adm}}
\newcommand{\pref}{\mathit{prf}}
\newcommand{\stage}{\mathit{stg}}
\newcommand{\semi}{\mathit{sem}}

\newcommand{\comp}{\mathit{com}}

\newcommand{\ground}{\mathit{grd}}
\newcommand{\rground}{\mathit{grd^*}}

\newcommand{\CRED}[1][]{{\sf Cred_{#1}}}
\newcommand{\SKEPT}[1][]{{\sf Skept_{#1}}}
\newcommand{\VER}[1][]{{\sf Ver_{#1}}}

\newcommand{\AF}{F}
\newcommand{\charF}{\mathcal{F}_\AF}
\renewcommand{\S}{\mathcal{S}}

\newcommand{\metasp}{{\tt{metasp}}}
\newcommand{\prefmetasp}{\mathit{\pref \_metasp}}
\newcommand{\stagemetasp}{\mathit{\stage \_metasp}}
\newcommand{\semimetasp}{\mathit{\semi \_metasp}}
\newcommand{\RminusBeta}{\mathrm{\att\_minus\_beta}}
\newcommand{\rgroundmetasp}{\mathit{\rground\_metasp}}
\newcommand{\resmetasp}{\mathit{res}}
\newcommand{\SplusR}{\mathit{range}}
\newcommand{\notSplus}{\mathrm{not\_in\_range}}

\begin{document}
\frontmatter          %

\mainmatter              %
\title{Making Use of Advances in Answer-Set 
Programming 
 for Abstract Argumentation Systems%
\thanks{%
Supported by the
Vienna Science and Technology Fund (WWTF) under
grant ICT08-028.}%
}
\author{Wolfgang Dvo\v{r}\'ak \and Sarah Alice Gaggl \and Johannes Wallner \and Stefan Woltran}
\authorrunning{Dvo\v{r}\'ak et al.} %
\institute{Institute of Information Systems, Database and Artificial Intelligence Group,\\
Vienna University of Technology, Favoritenstraße 9-11, 1040 Wien, Austria\\
EMail: \{dvorak, gaggl, wallner, woltran\}@dbai.tuwien.ac.at}

\maketitle              %

\begin{abstract}
Dung's famous 
abstract argumentation frameworks  
represent the core formalism for %
many problems and applications in 
the field of argumentation which significantly evolved within the last decade.
Recent work in the field has thus focused on implementations 
for these frameworks, 
whereby one 
of the main approaches
is to use Answer-Set Programming (ASP).
While some of the argumentation semantics can be nicely expressed 
within the ASP language, others required rather cumbersome
encoding techniques. %
Recent advances in ASP systems, in particular, the {\tt metasp}
optimization front-end for the ASP-package {\tt gringo/claspD} provides
direct commands to filter answer sets satisfying 
certain
subset-minimality (or -maximality) constraints.
This allows for much simpler encodings compared to the ones in 
standard ASP language.
In this paper, we experimentally compare 
the original encodings (for the argumentation semantics based on 
preferred, semi-stable, and respectively, stage extensions) %
with new {\tt metasp} encodings. %
Moreover, we provide novel encodings for the recently introduced 
resolution-based grounded semantics.
Our experimental results indicate that the {\tt metasp} approach works 
well in those cases where the complexity of the encoded problem is adequately
mirrored within the {\tt metasp} approach.
\keywords{Abstract Argumentation, Answer-Set Programming, Metasp}
\end{abstract}
\section{Introduction}
In Artificial Intelligence (AI), the area of argumentation
(the survey by Bench-Capon and Dunne~%
\cite{Bench-CaponD07} gives an excellent overview)
has become one of the central issues during the last decade.
Although there are now several branches within this area, there is 
a certain agreement that 
Dung's famous 
abstract argumentation frameworks (AFs) \cite{Dung95}
still represent the core formalism for many of the 
problems and applications in 
the field. %
In a nutshell,
AFs formalize statements together with a relation denoting rebuttals
between them, such that the semantics gives a handle to solve the
inherent conflicts between statements by selecting admissible subsets of them, 
but without taking the concrete contents of the statements into account. 
Several semantical principles how to select those subsets  have already
been proposed by Dung \cite{Dung95} but numerous other proposals have been 
made over the last years. %
In this paper we shall focus 
on the %
preferred \cite{Dung95}, semi-stable \cite{Caminada06}, stage \cite{Verheij96}, 
and the 
resolution-based grounded semantics \cite{BaroniDG11}.
Each of these semantics is based on some kind of $\subseteq$-maximality (resp. -minimality) and
thus is well amenable for the novel {\tt metasp} 
concepts which we describe below. 

Let us first talk about the general context of the paper, which is the realization of
abstract argumentation within the paradigm of Answer-Set Programming
(see \cite{Toni11} for an
overview%
).
We follow here 
the ASPARTIX%
\footnote{See \url{http://rull.dbai.tuwien.ac.at:8080/ASPARTIX} for a web front-end of ASPARTIX.}
approach~\cite{EglyGW10}, where a single program %
is used
to encode a particular argumentation semantics, while the instance of an argumentation framework
is given
as an input database. 
For problems located on the second level of the 
polynomial hierarchy (i.e.\ for preferred, stage, and semi-stable semantics)
ASP encodings turned out to be quite
complicated and hardly accessible for non-experts in ASP 
(we will sketch here the encoding for the stage semantics in some detail, 
since it has not been presented in \cite{EglyGW10}).
This is due to the
fact that tests for subset-maximality have to be done ``by hand'' in ASP 
requiring a certain saturation technique. 
However,
recent advances in ASP solvers, in particular, the {\tt metasp}
optimization front-end for the ASP-system {\tt gringo/claspD} allows
for much simpler encodings for such tests. More precisely, {\tt metasp} 
allows to use the traditional $\minimize$ statement (which in its standard
variant minimizes wrt.\ cardinality or weights, but not wrt.\ subset inclusion) also 
for selection among answer sets which are minimal (or maximal) wrt.\ subset inclusion
in certain predicates. 
Details about {\tt metasp}  can be found 
in~\cite{GebserKS11}.

Our first main contribution will be the practical comparison between handcrafted  encodings 
(i.e.\ encodings in the standard ASP language without the new
semantics for the $\minimize$ statement)
and  the much simpler 
{\tt metasp} 
encodings for argumentation semantics. 
The experiments show that the %
{\tt metasp} encodings 
do not necessarily result in longer runtimes. 
In fact, 
the {\tt metasp} encodings for the semantics located on the second level of the 
polynomial hierarchy
outperform the handcrafted 
saturation-based encodings. We thus can give additional evidence to
the observations in~\cite{GebserKS11}, where such a speed-up was
reported for encodings in a completely different application
area.

Our second contribution is the presentation of ASP encodings for the 
resolution-based grounded semantics~%
\cite{BaroniDG11}.
To the best of our knowledge, no implementation for this
quite interesting semantics has been released so far. In this paper, we
present a rather involved handcrafted encoding (basically following the $\NP$-algorithm presented in 
\cite{BaroniDG11}) but also two much simpler encodings (using {\tt metasp})
which rely on the original definition of the semantics. 
Our results indicate that {\tt metasp} is a very useful tool for 
problems known to be hard for the second-level, but one might loose performance in case 
{\tt metasp} is used for ``easier'' problems just for the sake of comfortability.
Nonetheless, we believe that the concept of the advanced $\minimize$ statement
is vital for ASP, since it allows for rapid prototyping of second-level encodings
without being an ASP guru. %

The remainder of the paper is organized as follows:
Section \ref{sec:back} 
provides the necessary background. %
Section \ref{sec:enc} then contains the ASP encodings 
for the semantics we are interested in here. We first discuss 
the handcrafted saturation-based encoding for stage semantics
(the ones for preferred and semi-stable are similar and already 
published). %
Then, in Section
\ref{sec:metasp} we provide  the novel 
{\tt metasp} encodings for all considered semantics. 
Afterwards,
in Section~\ref{sec:resbg}
we finally present an alternative encoding for
the resolution-based grounded semantics which better mirrors the complexity
of this semantics. Section~\ref{sec:exp} then presents our
experimental evaluation. We conclude the paper with a brief 
summary and discussion for future research directions.

\section{Background}\label{sec:back}
\subsection{Abstract Argumentation}

In this section we introduce (abstract) argumentation 
frameworks~\cite{Dung95} and
recall %
the 
semantics we study in this paper (see also
\cite{BaroniDG11,Baroni09}).
Moreover,
we highlight complexity results
for typical decision problems associated 
to such frameworks.

\begin{definition}\label{def:af}
An {\em argumentation framework (AF)} is a pair $\AF=(A,R)$ where $A$ 
is a set of arguments and $R \subseteq A \times A$ is the attack relation.
The pair 
$(a,b) \in R$ means that $a$ attacks %
$b$. 
An argument $a \in A$ is {\em defended} 
by a set $S \subseteq A$ 
if, for each $b \in A$ such that $(b,a) \in R$, 
there exists a $c \in S$ such that $(c,b) \in R$.
\end{definition}

\begin{example}\label{example:AF}
 Consider the AF $F=(A,R)$ with
    $A=\{a,b,c,d,e,f\}$ and
    $R=\{(a,b)$, $(b,d)$, $(c,b)$, $(c,d)$, $(c,e)$, $(d,c)$, $(d,e)$, $(e,f)\}$,
and the graph representation of $F$: 
\begin{center}
\begin{tikzpicture}[scale=1.4,>=stealth']
		\path 	node[arg](a){$a$}
			++(1,0) node[arg](b){$b$}
			++(1,.4) node[arg](c){$c$}
			++(0,-0.8) node[arg](d){$d$}
			++(1,0.4) node[arg](e){$e$}
			++(1,0) node[arg](f){$f$};
			;
		\path [left,->, thick]
			(a) edge (b)
			(c) edge (b)
			(b) edge (d)
			(d) edge (e)
			(c) edge (e)
			(e) edge (f)
			;
		\path [bend left, left, above,->, thick]
			(c) edge (d)
			(d) edge (c);
\end{tikzpicture}
\end{center}
\end{example}
Semantics for argumentation frameworks 
are given via a
function $\sigma$ which assigns 
to each AF $\AF=(A,R)$ 
a set
$\sigma(\AF)\subseteq 2^A$ of extensions. 
We shall consider here for 
$\sigma$ the functions
$\stable$, $\adm$, $\comp$, $\pref$,  
$\ground$, $\rground$, $\stage$, and $\semi$ which 
stand for 
stable, admissible, complete, preferred, grounded, resolution-based grounded,
stage, and  semi-stable semantics respectively.
Towards the definition of these semantics we have to introduce two 
more formal concepts.

\begin{definition}\label{def:charF}\label{def:range}
Given an  AF $\AF =(A,R)$.
The characteristic function $\charF: 2^A \Rightarrow 2^A$ of $\AF$ 
is defined as $\charF(S) = \{x \in A \mid x \mbox{ is defended by } S\}$.
Moreover, for a set $S\subseteq A$, we denote
the set of arguments attacked by $S$ as 
$S^\oplus_R=
\{ x\mid \exists y\in S \text{ such that } (y,x) \in R\}$, 
and define 
the \emph{range of $S$} as $S^+_R=S\cup S^\oplus_R$. 
\end{definition}

\begin{definition}\label{def:semantics}
Let $\AF=(A,R)$ be an AF.  A set $S\subseteq A$ is 
{\em conflict-free (in $\AF$)}, 
  if there are no %
$a, b \in S$, such that $(a,b) \in R$.
$\cf(\AF)$ denotes the collection of conflict-free sets of $\AF$.
For a conflict-free set $S \in \cf(\AF)$, it holds that
\begin{itemize}
\item  
$S\in\stable(\AF)$, 
if  $S^+_R = A$;
\item 
$S\in\adm(\AF)$, 
if  $S \subseteq \charF(S)$;
\item  
$S\in\comp(\AF)$, 
if $S=\charF(S)$;
\item 
$S\in\ground(\AF)$, 
if $S\in\comp(\AF)$ and 
there is no $T\in\comp(\AF)$ with $T\subset S$;
\item 
$S\in\pref(\AF)$, 
if $S\in\adm(\AF)$ and 
there is no $T\in\adm(\AF)$ with $T\supset S$;
\item 
$S\in\semi(\AF)$, if
$S\in\adm(\AF)$
and there is no 
$T\in\adm(\AF)$ with
$T^+_R\supset S^+_R$;
\item 
$S\in\stage(\AF)$,
if there is no $T\in\cf(\AF)$ in $\AF$, 
such that
$T^+_R\supset S^+_R$.
\end{itemize}
\end{definition}
We recall that for each AF $\AF$, 
the grounded semantics yields a unique extension, 
the grounded extension, which is the least fix-point of the characteristic function $\charF$.

\begin{example}\label{example:semantics}
Consider the AF $\AF$ from Example \ref{example:AF}. 
We have $\{a,d,f\}$ and $\{a,c,f\}$ as the stable extensions and thus %
$\stable(F)=\stage(F)=\semi(F)=\{\{a,d,f\},\{a,c,f\}\}$. 
The admissible sets of $F$ are
$\{\}$, $\{a\}$, $\{c\}$, $\{a,c\}$, $\{a,d\}$, $\{c,f\}$, $\{a,c,f\}$, $\{a,d,f\}$
and therefore  $\pref(F)=\{\{a,c,f\}$,$\{a,d,f\}\}$.
Finally we have  $\comp(F)=\{\{a\}$, $\{a,c,f\}$, $\{a,d,f\}\}$, with $\{a\}$ being the grounded extension.
\end{example}
On the base of these semantics one can define the family of resolution-based semantics~\cite{BaroniDG11},
with the resolution-based grounded semantics being the most popular instance.
\begin{definition}\label{def:resbased}
A resolution $\beta \subset R$ of an $\AF=(A,R)$
contains exactly one of the attacks $(a,b)$, $(b,a)$ if $\{(a,b), (b,a)\} \subseteq R$, $a\neq b$,
and no further attacks.
 A set $S\subseteq A$ is a resolution-based grounded extension of $F$ if 
 (i) there exists a resolution $\beta$ such that $S=\ground((A,R \setminus\beta))$%
;\footnote{Abusing notation slightly, we use $\ground(\AF)$ for denoting the unique grounded extension of $\AF$.}
 and (ii) there is no resolution $\beta'$  such that $\ground((A,R \setminus\beta')) \subset S$.
\end{definition}

\begin{example}\label{example:resbased}
Recall the AF $F=(A,F)$ from Example \ref{example:AF}. There is one mutual attack and thus we have two 
resolutions $\beta_1=\{(c,d)\}$ and $\beta_2=\{(d,c)\}$.
Definition \ref{def:resbased} gives us two candidates, 
namely 
$\ground((A,R \setminus \beta_1))=
\{a,d,f\}$ and 
$\ground((A,R \setminus \beta_2))=
\{a,c,f\}$;
as they are not in $\subset$-relation they are %
the resolution-based grounded extensions
of $F$.
\end{example}
\noindent
We now turn to the complexity of reasoning in AFs. 
To this end, we define 
the following decision problems for the semantics $\sigma$ introduced in 
Definitions~\ref{def:semantics} and \ref{def:resbased}:
\vspace{-1pt}
\begin{itemize}
\item \emph{Credulous Acceptance} $\CRED[\sigma]$: Given AF $\AF=(A,R)$ and an argument $a\in A$. Is $a$ contained in some $S\in \sigma(\AF)$?
\item \emph{Skeptical Acceptance} $\SKEPT[\sigma]$: Given AF $\AF=(A,R)$ and an argument $a\in A$. Is $a$ contained in each $S\in \sigma(\AF)$?
\item \emph{Verification of an extension} $\VER[\sigma]$: Given AF $\AF=(A,R)$ and a set of arguments $S \subseteq A$. Is $S\in \sigma(\AF)$?
\end{itemize}
\vspace{-1pt}
\noindent
We assume the reader has knowledge about standard complexity classes like $\Pol$ and $\NP$
and recall that $\SigmaP{2}$
is the class of decision problems that can be decided in polynomial time
using a nondeterministic Turing machine with access to an $\NP$-oracle.
The class $\PiP{2}$ is defined as the complementary class of $\SigmaP{2}$, i.e. $\PiP{2}=\co\SigmaP{2}$.

In Table~\ref{tab:compl} we summarize complexity results relevant for our work \cite{BaroniDG11,DimopoulosT96,DunneB02,CaminadaDunne08,DvorakW10}.

\begin{table}[t!]
\setlength{\tabcolsep}{0.6em}
\renewcommand{\arraystretch}{1.5}
\begin{center}
\begin{tabular}{l|cccc}
& $\pref$ & $\semi$ & $\stage$ &  $\rground$\\
\hline
$\CRED[\sigma]$ & $\NP$-c & $\SigmaP{2}$-c & $\SigmaP{2}$-c & $\NP$-c\\
$\SKEPT[\sigma]$ & $\PiP{2}$-c & $\PiP{2}$-c & $\PiP{2}$-c & $\coNP$-c\\ 
$\VER[\sigma]$ & $\coNP$-c & $\coNP$-c & $\coNP$-c & in $\Pol$\\
\end{tabular}
\end{center}
\renewcommand{\arraystretch}{1.0}
\caption{Complexity of abstract argumentation (${\cal C}$-c denotes completeness for class ${\cal C}$)}
\label{tab:compl}
\end{table}

\subsection{Answer-Set Programming}
We first give a brief overview of the syntax and
semantics of disjunctive logic programs under the answer-sets semantics
\cite{GelfondL91}; for further background, see
\cite{%
dlv}.

We fix a countable set $\U$ of {\em (domain) elements}, also called \emph{constants};
and suppose a total order $<$ over the domain elements.
An {\em atom} is an expression
$p(t_{1},\ldots,t_{n})$, where $p$ is a {\em predicate} of arity $n\geq 0$
and each $t_{i}$ is either a variable or an element from $\U$.
An atom is \emph{ground} if it is free of variables.
$\BU$ denotes the set of all ground atoms over $\U$.

A \emph{(disjunctive) rule} $r$ is of the form
\begin{center}
$a_1\ \vee\ \cdots\ \vee\ a_n\ \la
        b_1,\ldots, b_k,\
        \naf b_{k+1},\ldots,\ \naf b_m$,
\end{center}
with $n\geq 0,$ $m\geq k\geq 0$, $n+m > 0$, where
$a_1,\ldots ,a_n,b_1,\ldots ,b_m$ are
atoms, and ``$\naf$'' stands for {\em default negation}.
The \emph{head} of $r$ is the set
$\head{r}$ = $\{a_1\commadots a_n\}$ and
the \emph{body} of $r$ is
$\body{r}=
\{b_1,\ldots, b_k,$ $\naf b_{k+1},\ldots,$ $\naf  b_m\}$.
Furthermore, $\bodyp{r}$ = $\{b_{1}\commadots b_{k}\}$ and
$\bodyn{r}$ = $\{b_{k+1}\commadots b_m\}$.
A rule $r$ is \emph{normal} if $n \leq 1$ and a %
\emph{constraint} if $n=0$. 
A rule $r$ is \emph{safe} if each variable in $r$ occurs in $\bodyp{r}$.
A rule $r$ is \emph{ground} if no variable occurs in $r$.
A \emph{fact} is a ground rule without disjunction and empty body. 
An \emph{(input) database} is a set of facts.
A program is a finite set of disjunctive rules. 
For a program $\p$ and an input database $D$, we often write $\p{(D)}$ instead of $D\cup\p$.
If each rule in a program is 
normal (resp.\ ground), we call the program normal (resp.\ ground).
Besides disjunctive and normal program, we consider here
the class of optimization programs, i.e. normal programs 
which additionally contain 
$\minimize$ statements
\begin{equation}\label{min}
\minimize[l_1=w_1@J_1,\dots ,l_k=w_k@J_k], 
\end{equation}
where $l_i$ is a literal, $w_i$ an integer weight and $J_i$ an integer priority level.%

For any program \p{}, let \UP{}
be the set of all constants appearing in \p{}.
$\GP$ is  the set of rules
$r\sigma$
obtained by applying, to each rule 
$r\in\p{}$, all possible
substitutions $\sigma$ from the variables
in $r$ to elements of $\UP{}$.
An \emph{interpretation} $I\subseteq \BU$ 
\emph{satisfies} %
a ground rule $r$
iff $\head{r} \cap I \neq \emptyset$ whenever
$\bodyp{r}\subseteq I$ and $\bodyn{r} \cap I = \emptyset$.
$I$ satisfies a ground program $\p$,
if each $r\in\p$
is satisfied by $I$.
A non-ground rule $r$ (resp., a program $\p$)
is satisfied by an interpretation $I$ iff
$I$ satisfies all groundings of $r$ (resp., $\GP$).
$I \subseteq \BU$ is an \emph{answer set}
of $\p$
iff it is a subset-minimal set
satisfying
the \emph{Gelfond-Lifschitz reduct}
$
\p^I=\{ \head{r} \derives \bodyp{r} \mid I\cap
\bodyn{r} = \emptyset, r \in \GP\}
$.
For a program $\p$,
we denote the set of its %
answer sets by 
$\AS(\p)$.

For semantics of optimization programs, 
we interpret 
the $\minimize$ statement wrt.\ subset-inclusion:
For any sets $X$ and $Y$ of atoms, we have $Y\subseteq^w_J X$, if for any weighted literal $l=w@J$ occurring in (\ref{min}), $Y\models l$ implies $X\models l$. Then, $M$ is a collection of relations of the form $\subseteq^w_J$ for priority levels $J$ and weights $w$. 
A standard answer set (i.e.\ not taking the minimize statements into account) $Y$ of $\p$ \emph{dominates} a standard answer set $X$ of $\p$ wrt.\ $M$ if there are a priority level $J$ and a weight $w$ such that $X\subseteq^w_J Y$ does not hold for $\subseteq^w_J\in M$, while $Y\subseteq^{w'}_{J'} X$ holds for all $\subseteq^{w'}_{J'}\in M$ where $J'\geq J$.
Finally a standard answer set $X$ is an answer set of an optimization program $\p$ wrt.\ $M$ if there is no standard answer set $Y$ of $\p$ that dominates $X$ wrt.\ $M$.

Credulous and skeptical reasoning in terms of programs is defined as follows. 
Given a program $\p$ and a set of ground atoms $A$. Then, 
we write $\p\models_c A$ (credulous reasoning), if $A$ is contained in some answer set of $\p$;
we write $\p\models_s A$ (skeptical reasoning), if $A$ is contained in each answer set of $\p$.

We briefly recall some complexity results for disjunctive logic programs.
In fact, since we will deal with fixed programs we focus on results for data complexity.
Depending on the concrete definition of $\models$, we give the complexity
results in Table~\ref{tab:1} 
(cf.~\cite{dant-etal-01} and the references therein).
We note here, that even %
normal programs together with the optimization technique 
have a worst case complexity of $\SigmaP{2}$ (resp. $\PiP{2}$).
\begin{table}[t]
\setlength{\tabcolsep}{0.6em}
\renewcommand{\arraystretch}{1.5}
\begin{center}
\begin{tabular}{l|ccccc}
$e$ 
       &  %
normal programs &       disjunctive program & optimization programs \\
\hline
$\models_c$ &  %
                $\NP$ &         $\SigmaP{2}$ & $\SigmaP{2}$\\
$\models_s$ &  %
       $\CONP$ &       $\PiP{2}$ & $\PiP{2}$\\
\end{tabular}
\end{center}
\caption{Data Complexity for logic programs (all results are completeness results).}
\label{tab:1}
\end{table}
Inspecting Table~\ref{tab:compl} one can see which kind of encoding is appropriate for an argumentation semantics.

\section{Encodings of AF Semantics}\label{sec:enc}
In this section we first show how to represent AFs in ASP and we discuss three programs which we need later on in this section\footnote{We make use of some program modules already defined in \cite{EglyGW10}.}.
Then, in Subsection~\ref{sec:satenc} we exemplify on the stage semantics the saturation technique for encodings which solve associated problems which are on the second level of the polynomial hierarchy. 
In Subsection~\ref{sec:metasp} 
we will make use of the newly developed \metasp{} optimization technique.
In Subsection~\ref{sec:resbg} we give an alternative encoding based on the algorithm of Baroni {\em et al.} in~\cite{BaroniDG11}, which respects the lower complexity of resolution-based grounded semantics.

All our programs are fixed which means that the only translation required, is to give an AF $F$ as input database $\hat{F}$ to the program $\pi_\sigma$ for a semantics $\sigma$.
In fact, for an AF $F=(A,R)$, we define $\hat{F}$ as 
\[
\hat{F} = \{ \; \arg(a) \mid a\in A\} \cup \{\att(a,b)\mid (a,b)\in
R  \; \}.
\] 
In what follows,
we use unary predicates 
$\inA/1$ and $\out/1$ to perform a guess for a set $S\subseteq A$, 
where $\inA(a)$ represents that $a\in S$. 
The following notion of correspondence is relevant for our purposes.

\begin{definition}
Let $\S\subseteq 2^\U$ be a collection of sets of domain elements and 
let $\I\subseteq 2^\BU$ be a collection of sets of ground atoms.
We say that $\S$ and $\I$ correspond to each other, in symbols
$\S\corr \I$, iff 
(i) for each $S\in \S$, there exists an $I\in\I$, such that
$\{ a \mid \inA(a)\in I\} = S$;
(ii) for each $I\in \I$, it holds that 
$\{ a \mid \inA(a)\in I\}\in \S$; and
(iii) $|\S| = |\I|$.
\label{def:corr}
\end{definition}
Consider an AF $F$.
The following program
fragment guesses, when augmented by $\hat{F}$, any subset $S\subseteq
A$ and then checks whether the guess is conflict-free in $F$:

\begin{eqnarray*}
\pi_\cf & = \{ & 
        \inA(X) \la \naf \out(X), \arg(X);\; \\
        &&\out(X) \la \naf \inA(X), \arg(X);\;\\
        && \la \inA(X), \inA(Y), \att(X,Y) \; \}.
\end{eqnarray*}
\begin{proposition}
 For any AF $F$, $\cf(F)\corr \AS(\pi_\cf(\hat{F})).$
\end{proposition}
Sometimes we have to avoid the use of negation. 
This might either be the case for the saturation technique 
or if a simple program can be solved without a Guess\&Check 
approach. Then, encodings typically rely on a form of loops where
all domain elements are visited and it is checked whether a desired
property holds for all elements visited so far. We will use this
technique in our saturation-based encoding in the upcoming subsection, 
but also for %
computing
the grounded extension in 
Subsection~\ref{sec:metasp}. 
For this purpose the program 
$\pi_\order$, which is taken from \cite{EglyGW10}, is used to encode the infimum, successor and supremum of an order $<$ over the domain elements in the predicates $\inf/1, \succ/2$ and $\sup/1$ respectively. The order over the domain elements is usually provided by common ASP solvers.

Finally, the following module computes for a guessed subset $S
\subseteq A$ the range $S^+_R$ (see Def. \ref{def:range}) of $S$ in an
AF $(A,R)$.

\begin{eqnarray*}
 \pi_\SplusR & = & \{ 
	\inSplus(X) \la \inA(X); \\
&&
\hphantom{\{}
	\inSplus(X) \la \inA(Y), \att(Y,X);\\ 
&&	
\hphantom{\{}
\notSplus(X) \la \arg(X), \naf \inSplus(X) \}.
\end{eqnarray*}
\subsection{Saturation Encodings}\label{sec:satenc}
In this subsection we make use of the saturation technique introduced by Eiter and Gottlob in~\cite{EiterG95}. 
In~\cite{EglyGW10}, 
this technique was already used to encode the preferred and semi-stable semantics.
Here we give the encodings for the stage semantics, 
which is %
similar to the one of semi-stable semantics,
to exemplify the use of the saturation technique. 

In fact, for an AF $F=(A,R)$ and $S\in \cf(F)$ we need to check whether no $T\in\cf(F)$ with $S^+_R\subset T^+_R$ exists. Therefore we have to guess an arbitrary set $T$ and saturate in case 
(i) $T$ is not conflict-free, and
(ii) $S^+_R\not\subset T^+_R$.
Together with 
$\pi_\cf$ this is done with the following module, where
$\inA/1$ holds the current guess for $S$ and 
$\inN/1$ holds the current guess for $T$.
More specifically, 
rule $
\spoil                  \la \inN(X),\inN(Y),\att(X,Y)$
checks for (i) and the remaining two rules with $\spoil$ in the head
fire in case
$S^+_R= T^+_R$ (indicated by predicate $\eqplus/0$ described below),
or
there exists an $a\in S^+_R$ such that 
$a\notin T^+_R$ 
(here we use predicate 
$\inSplus/1$
from above and predicate
$\notdefeated/1$ which we also present 
below).
As is easily checked one of these two conditions
holds exactly if (ii) holds.
\begin{eqnarray*}
\pi_\pispoilstage & =\{ & 
        \inN(X) \vee \outN(X) \la \arg(X); \\ %
&&      \spoil                  \la \inN(X),\inN(Y),\att(X,Y);\\ %
&&      \spoil                  \la \eqplus;\\ %
&& \spoil \la \inSplus(X), \notdefeated(X);\\
&&      \inN(X)                 \la \spoil, \arg(X); \\
&&      \outN(X)                \la \spoil, \arg(X);\\ %
&&                          \la \naf \spoil \; \}. %
\end{eqnarray*}

\noindent
For the definition of predicates 
$\notdefeated/1$ and 
$\eqplus/0$ 
we make use of the aforementioned loop technique and predicates from program 
$\pi_\order$. %
\begin{eqnarray*}
\pi_\rangeN & = \{ &      \notdefeatedupto(X,Y)   \la \inf(Y), \outN(X), \outN(Y); \\
&&      \notdefeatedupto(X,Y)   \la \inf(Y), \outN(X), \naf \defeat(Y,X); \\
&&      \notdefeatedupto(X,Y)   \la \succ(Z,Y), \notdefeatedupto(X,Z), \outN(Y);\\
&&      \notdefeatedupto(X,Y)   \la \succ(Z,Y), \notdefeatedupto(X,Z),\\
&& \hphantom{\notdefeatedupto(X,Y)   \la} \naf \defeat(Y,X);\\
&&      \notdefeated(X)         \la \sup(Y), \outN(X), \notdefeatedupto(X,Y);\\
&&	  \inRangeN(X) \la \inN(X); \\
&&	\inRangeN(X) \la \outN(X), \inN(Y), \defeat(Y,X) \; \}.
\end{eqnarray*}

\begin{eqnarray*}
\pi_\equalplus & =   \{ & 
        \eqplusupto(X) \la \inf(X), \inSplus(X), \inRangeN(X);  \\
&&      \eqplusupto(X) \la \inf(X), \notSplus(X), \notdefeated(X); \\
&&      \eqplusupto(X) \la \succ(Z,X), \inSplus(X), \inRangeN(X), \eqplusupto(Z);\\
&&      \eqplusupto(X) \la \succ(Y,X), \notSplus(X), \notdefeated(X),\eqplusupto(Y);\\
&&      \eqplus \la \sup(X), \eqplusupto(X) \; \};
\\
\end{eqnarray*}
\begin{proposition}
 For any AF $F$, $\stage(F)\corr \AS(\pi_\stage(\hat{F}))$, where
$\pi_\stage  =  \pi_\cf \cup \pi_\order \cup \pi_\SplusR \cup \pi_\rangeN \cup \pi_\equalplus
\cup \pi_\pispoilstage$.
\end{proposition}

\subsection{Meta ASP Encodings}\label{sec:metasp}
The following encodings for preferred, semi-stable and stage semantics are written using the 
$\minimize[\cdot]$ statement when evaluated with 
the subset minimization semantics provided by
\metasp{}.
For our encodings we do not need prioritization and weights, therefore these are omitted (i.e.\ set to default) in the minimization statements. The fact {\tt optimize(1,1,incl)} is added to the meta ASP encodings, to indicate that we use subset inclusion for the optimization technique using priority and weight $1$.

We now look at the encodings for the preferred, semi-stable and stage
semantics using this minimization technique. First we need one
auxiliary module for admissible extensions. 

\begin{eqnarray*}
\pi_\adm & = & 
\pi_\cf\cup \{
                \defeated(X) \la \inA(Y), \defeat(Y,X);\; \\
                && 
\hphantom{ \pi_\cf\cup \{}
\la \inA(X), \defeat(Y,X), \naf \defeated(Y) \}.
\end{eqnarray*}
\vspace{-4pt}

\noindent
Now the modules for preferred, semi-stable and stage semantics are easy to encode using the minimization statement of \metasp{}. For the preferred semantics we take the module $\pi_\adm$ and minimize the $\out/1$ predicate. This in turn gives us the subset-maximal admissible extensions, which captures the definition of preferred semantics. The encodings for the semi-stable and stage semantics are similar. Here we minimize the predicate $\notSplus/1$ from the $\pi_\SplusR$ module. 

\begin{eqnarray*}
\pi_\prefmetasp & = & \pi_\adm \cup \{ \minimize[\out]\}. \\
\pi_\semimetasp & = & 
\pi_\adm 
\cup 
\pi_\SplusR 
\cup \{ \minimize[\notSplus] \}. \\
\pi_\stagemetasp & = & 
\pi_\cf 
\cup 
\pi_\SplusR 
\cup \{ \minimize[\notSplus] \}.
\end{eqnarray*}
\vspace{-4pt}

\noindent
The following results follow now quite directly.

\begin{proposition}
For any AF $F$, we %
have
\begin{enumerate}
\item 
$\pref(F)\corr \AS(\pi_\prefmetasp(\hat{F}))$,
\item 
$\semi(F)\corr  \AS(\pi_\semimetasp(\hat{F}))$, and
\item 
$\stage(F)\corr \AS(\pi_\stagemetasp(\hat{F}))$.
\end{enumerate}
\end{proposition}
Next we give two different encodings for computing resolution-based grounded extensions.
Both encodings use subset minimization for the resolution part, i.e.\ the 
resulting extension is subset minimal with respect to all possible resolutions.
The first one computes the grounded extension for the guessed resolution explicitly
(%
adapting the encoding from
\cite{EglyGW10};
instead of the $\att$ predicate we use 
$\RminusBeta$, since we need the grounded extensions of a restricted $\att$ relation). 
In fact,
the $\pi_\resmetasp$ module which we give next guesses this restricted $\att$ relation $\{R \setminus \beta\}$ for a resolution $\beta$.

\begin{eqnarray*}
\pi_\resmetasp & = \{ &
	\RminusBeta(X,Y) \la \att(X,Y), \naf \RminusBeta(Y,X),\\
&&	\hphantom{\RminusBeta(X,Y) \la\ } X \not = Y;\\
&&	\RminusBeta(X,Y) \la \att(X,Y), \naf \att(Y,X); \\
&&	\RminusBeta(X,X) \la \att(X,X) \}.
\end{eqnarray*}

\noindent
The second encoding uses the \metasp{} subset minimization 
additionally to get the grounded extension from the 
complete extensions of the current resolution (recall that the 
grounded extension is in fact the unique subset-minimal complete
extension). We again use the restricted $\att$ relation.

\begin{eqnarray*}
\pi_\comp & =  & \pi_\adm 
\cup  \{\;
\notdefended(X) \la \RminusBeta(Y,X), \naf \defeated(Y);\\
 && 
\hphantom{\pi_\adm\cup  \{}
\la \out(X), \naf \notdefended(X)    \; \}.\\\
\end{eqnarray*}
\noindent
Now we can give the two encodings for resolution-based grounded
semantics.

\begin{eqnarray*}
\pi_\rgroundmetasp & = & \pi_\ground \cup \pi_\resmetasp \cup \{ \minimize[\inA] \}\\
\pi_\rgroundmetasp' & = & \pi_\comp \cup \pi_\resmetasp \cup \{ \minimize[\inA] \}. 
\end{eqnarray*}
\begin{proposition}
For any AF $F$ and
 $\pi \in \{\pi_\rgroundmetasp,\pi'_\rgroundmetasp\}$, %
	$\ground^*(F)$ corresponds to $\AS(\pi(\hat{F}))$ in the sense of Definition \ref{def:corr}, but without property (iii).
\end{proposition}

\subsection{Alternative Encodings for Resolution-based Grounded Semantics}\label{sec:resbg}

So far, we have shown two encodings for the 
resolution-based grounded semantics 
via optimization programs, i.e.\ 
we made use of the $\minimize$ statement under the 
subset-inclusion semantics.
From the complexity point of view this is not adequate, 
since we expressed a problem on the $\NP$-layer 
(see Table~\ref{tab:compl}) via an encoding
which implicitly makes use of disjunction (see Table~\ref{tab:1} 
for the actual complexity of optimization programs).
Hence, we provide here an alternative encoding for the 
resolution-based grounded semantics based on the verification algorithm proposed by Baroni {\em et al.} in~\cite{BaroniDG11}. This encoding 
is just a normal program and thus located at the right level of complexity.

We need some further notation. For an AF $F=(A,R)$ and a set $S\subseteq A$ we define $F|_{S}=((A\cap S),R\cap(S\times S))$ as the \emph{sub-framework} of $F$ wrt $S$; furthermore we also use $F-S$ as a shorthand for $F|_{A\setminus S}$. %
By $\sccs$,  we denote the set of strongly connected components of 
an AF $F=(A,R)$ which identify the vertices of a maximal strongly connected\footnote{A directed graph is called \emph{ strongly connected} if there is a directed path from each vertex in the graph to every other vertex of the graph.} subgraphs of $F$; $\sccs$ is thus a partition of $A$. A partial order $\prec_F$ over $\sccs=\{C_1,\dots ,C_n\}$, denoted as $(C_i\prec_F C_j)$ for $i\neq j$, is defined, if $\exists x\in C_i, y\in C_j$ such that there is a directed path from $x$ to $y$ in $F$. 
\begin{definition}\label{def:mr}
A $C\in\sccs$ is {\em minimal relevant} (in an AF $F$) iff $C$ is a minimal element of $\prec_F$ and $F|_C$ satisfies the following: %
\begin{enumerate}
  \item[(a)] the attack relation $R(F|_C)$ of $F$ is irreflexive,
    i.e.\ $(x,x)\not\in R(F|_C)$ for all arguments $x$; 
  \item[(b)] $R(F|_C)$ is symmetric, i.e. $(x,y)\in R(F|_C) \Leftrightarrow (y,x)\in R(F|_C)$;
  \item[(c)] the {\em undirected graph} %
obtained by replacing each (directed) pair $\{(x,y),(y,x)\}$ in $F|_C$
with a single undirected edge $\{x,y\}$ is acyclic. 
\end{enumerate}
The set of minimal relevant SCCs in $F$ is denoted by $\pmr(F)$.
\end{definition}

\begin{proposition}[\cite{BaroniDG11}]\label{prop:rbg} 
Given an AF $F=(A,R)$ such that $(F-S^+_R)\neq (\emptyset, \emptyset)$ and $\pmr(F-S^+_R)\neq\emptyset$, where $S = \ground(F)$, 
a set 
$U\subseteq A$ 
of arguments 
is {\em resolution-based grounded} in $F$, i.e. $U\in \rground(F)$ iff the following conditions 
hold:
\begin{enumerate}
 \item[(i)] $U\cap S^+_R =S$; %
 \item[(ii)] $(T\cap \Pi_F) \in \stable(F|_{\Pi_F})$, where
$T=U\setminus S^+_R$, and $\Pi_F=\bigcup_{V\in \pmr(F-S^+_R)}V$; 
 \item[(iii)] $(T\cap \Pi_F^C)\in \rground(F|_{\Pi_F^C} -(S^+_R \cup (T\cap\Pi_F)^\oplus_R))$, where $T$ and $\Pi_F$ are as in (ii) and 
$\Pi_F^C = A\setminus \Pi_F$.
\end{enumerate}
\end{proposition}

\noindent
To illustrate the conditions of Proposition~\ref{prop:rbg}, let us have a look at our example.
\begin{example}
 Consider the AF $F$ of Example~\ref{example:AF}.
Let us check whether $U=\{a,d,f\}$
is resolution-based grounded in $F$, i.e.\
whether $U\in\rground(F)$. 
$S=\{a\}$ is the grounded extension of $F$ and $S^+_R=\{a,b\}$, hence the first Condition (i) is satisfied. 
We obtain $T=\{d,f\}$ and $\Pi_F=\{c,d\}$. We observe that $T\cap \Pi_F=\{d\}$ is a stable
extension of the AF $F|_{\Pi_F}$; that satisfies Condition (ii). 
Now we need to check Condition (iii); we first identify the necessary sets:
$\Pi_F^C=\{a,b,e,f\}$, $T\cap \Pi_F^C=\{f\}$ and $(T\cap\Pi_F)^\oplus_R=\{c,e\}$. 
It remains to check $\{f\}\in \rground(\{f\},\emptyset)$ which is easy to see. Hence, $U\in\rground(F)$.
\end{example}
The following encoding is based on the Guess\&Check procedure which was 
also used for the encodings in \cite{EglyGW10}.
 After guessing all conflict-free sets  with the program $\pi_\cf$, we check whether the conditions of Definition~\ref{def:mr} and Proposition~\ref{prop:rbg} hold. Therefore the program $\pi_\pargset$ makes a copy of the actual arguments, defeats and the guessed set to the predicates $\argset/2, \defeatN/3$ and $\inU/2$. The first variable in these three predicates serves as an identifier for the iteration of the algorithm (this is necessary to handle the recursive nature of 
Proposition~\ref{prop:rbg}).
In all following predicates we will use the first variable of each predicate like this. 
As in some previous encodings in this paper, 
we use the program $\pi_\order$ to obtain an order over the arguments, and we start our computation with the infimum represented by the predicate $\inf/1$.
\begin{eqnarray*}
 \pi_\pargset &=\{&
	\argset(N,X) \la \arg(X), \inf(N);\\
	&&\inU(N,X) \la \inA(X), \inf(N);\\
	&&\defeatN(N,Y,X) \la \argset(N,X), \argset(N,Y), \att(Y,X)\; \}.
\end{eqnarray*}
We use here the program $\pi_\pdefended$ (which is a slight variant of the program $\pi_\mathit{defended}$) together with the program $\pi_\pgroundN$ where we perform a fixed-point computation of the predicate $\defended/2$, but now we use an additional argument $N$ for the iteration step where predicates $\argset/2$, $\defeatN/3$ and $\inS/2$ replace $\arg/1$, $\defeat/2$ and $\inA/1$. 
In $\pi_\pgroundN$ we then obtain the predicate $\inS(N,X)$ which identifies argument $X$ to be in the grounded extension of the iteration $N$.

\begin{eqnarray*}
 \pi_\pgroundN & = \pi_\cf \cup \pi_\order \cup \pi_\pargset \cup \pi_\pdefended \cup \{& \inS(N,X) \la \defended(N,X)\; \}.
\end{eqnarray*}
The next module $\pi_\pcut$ computes the arguments in $(F-S^+_R)$, represented by the predicate $\cut/2$, via predicates $\alphaN/2$ and $\uAlpha/2$ (for $S^+_R$ and $U\cap S^+_R$). The two constraints 
check condition (i) of Proposition~\ref{prop:rbg}.
\begin{eqnarray*}
 \pi_\pcut &=\{&
	\alphaN(N,X) \la \inS(N,X); \\
	&&\alphaN(N,X) \la \inS(N,Y), \defeatN(N,Y,X); \\
	&&\uAlpha(N,X) \la \inU(N,X), \alphaN(N,X); \\
	&&\la \uAlpha(N,X), \naf \inS(N,X); \\
	&&\la \naf \uAlpha(N,X), \inS(N,X); \\
	&&\cut(N,X) \la \argset(N,X), \naf \alphaN(N,X)	\; \}.
\end{eqnarray*}
The module $\pi_\pmr$ computes $\Pi_F=\bigcup_{V\in \pmr(F-S^+_R)}V$, where $\mr(N,X)$ denotes that an argument is contained in a set $V\in\pmr$.
Therefore we need to check all three conditions of Definition~\ref{def:mr}. 
The first two rules compute the predicate $\reach(N,X,Y)$ if there is a path between the arguments $X,Y\in (F-S^+_R)$. 
With this predicate we will identify the SCCs. The third rule computes $\nmr/2$ for all arguments violating Condition (a). 
Next we need to check Condition (b). 
With $\nsym/2$ we obtain those arguments which do not have a symmetric attack to any other argument from the same component.
Condition (c) is a bit more tricky. With predicate $\reachnotvia/4$ 
we say that there is a path from $X$ to $Y$ not going over argument $V$ in the framework $(F-S^+_R)$. 
With this predicate at hand we can check for cycles with $\cyc/4$. 
Then, to complete Condition (c) we derive $\bad/2$ for all arguments which are connected to a cycle (or a self-defeating argument). 
In the predicate $\posmr/2$, we put all the three conditions together and say that 
an argument $x$ is possibly in a set  $V\in\pmr$ if  
(i) $x\in (F-S^+_R)$, 
(ii) $x$ is neither connected to a cycle nor self-defeating, and 
(iii) for all $y$ it holds that $(x,y)\in (F-S^+_R)\Leftrightarrow (y,x)\in (F-S^+_R)$. 
Finally we only need to check if the SCC obtained with $\posmr/2$ is a minimal element of $\prec_F$. 
Hence we get with $\notminimal/2$ all arguments not fulfilling this, 
and in the last rule we obtain with $\mr/2$ the arguments contained in a minimal relevant SCC.

\begin{eqnarray*}
 \pi_\pmr\! &=\!\{&
 	\reach(N,X,Y) \la \cut(N,X), \cut(N,Y), \defeatN(N,X,Y); \\
	&&\reach(N,X,Y) \la \cut(N,X), \defeatN(N,X,Z), \reach(N,Z,Y),\\
&&\hphantom{\reach(N,X,Y) \la\ } X!=Y; \\
	&&\nmr(N,X) \la \cut(N,X), \defeatN(N,X,X); \\ 
	&&\nsym(N,X) \la \cut(N,X), \cut(N,Y), \defeatN(N,X,Y), \\
&&\hphantom{\nsym(N,X) \la\ }\naf \defeatN(N,Y,X), \reach(N,X,Y), \reach(N,Y,X), X!=Y; \\ 
	&&\nsym(N,Y) \la \cut(N,X), \cut(N,Y), \defeatN(N,X,Y), \\
 &&\hphantom{\nsym(N,X) \la\ }\naf \defeatN(N,Y,X), \reach(N,X,Y),\reach(N,Y,X), X!=Y; \\ 
	&&\reachnotvia(N,X,V,Y) \la \defeatN(N,X,Y), \cut(N,V),  \\
&&\hphantom{\reachnotvia(N,X,V,Y) \la\ }\reach(N,X,Y),\reach(N,Y,X), X!=V,Y!=V; \\
	&&\reachnotvia(N,X,V,Y) \la \reachnotvia(N,X,V,Z),  \reach(N,X,Y),\\
&&\hphantom{\reachnotvia(N,X,V,Y) \la\ } \reachnotvia(N,Z,V,Y),\reach(N,Y,X),\\
&&\hphantom{\reachnotvia(N,X,V,Y) \la\ } Z!=V, X!=V,Y!=V; \\
	&&\cyc(N,X,Y,Z) \la \defeatN(N,X,Y), \defeatN(N,Y,X),  \\
&&\hphantom{\cyc(N,X,Y,Z) \la\ }\defeatN(N,Y,Z), \defeatN(N,Z,Y),\\
&&\hphantom{\cyc(N,X,Y,Z) \la\ } \reachnotvia(N,X,Y,Z), X!=Y, Y!=Z, X!=Z; \\ 
	&&\bad(N,Y) \la \cyc(N,X,U,V), \reach(N,X,Y), \reach(N,Y,X); \\
	&&\bad(N,Y) \la \nmr(N,X), \reach(N,X,Y), \reach(N,Y,X); \\
	&&\posmr(N,X) \la \cut(N,X), \naf \bad(N,X), \naf \nmr(N,X),\\
&&\hphantom{\posmr(N,X) \la\ } \naf \nsym(N,X); \\
	&&\notminimal(N,Z) \la \reach(N,X,Y), \reach(N,Y,X),\\
&&\hphantom{\notminimal(N,Z) \la\ } \reach(N,X,Z), \naf \reach(N,Z,X); \\
	&&\mr(N,X) \la \posmr(N,X), \naf \notminimal(N,X)\; \}.
\end{eqnarray*}
We now turn to Condition (ii) of Proposition~\ref{prop:rbg}, where the first rule in $\pi_\pstableN$ computes the set $T=U\setminus S^+_R$. Then we check whether $T=\emptyset$ and $\pmr(F-S^+_R)=\emptyset$ via predicates $\emptyT/1$ and $\notmr/1$. If this is so, we terminate the iteration in the last module $\pi_\piterate$. The first constraint eliminates those guesses where $\pmr(F-S^+_R)=\emptyset$ but $T\neq\emptyset$, because the algorithm is only defined for AFs fulfilling this. Finally we derive the arguments which are defeated by the set $T$ in the $\pmr$ denoted by $\defeated/2$, and with the last constraint we eliminate those guesses where there is an argument not contained in $T$ and not defeated by $T$ in $\pmr$ and hence $(T\cap\Pi_F)\not\in \stable(F|_{\Pi_F})$.

\begin{eqnarray*}
 \pi_\pstableN &=\{&
	{\mathrm t}(N,X) \la \inU(N,X), \naf \inS(N,X); \\
	&&\nemptyT(N) \la {\mathrm t}(N,X); \\
	&&\emptyT(N) \la \naf \nemptyT(N), \argset(N,X); \\
	&&\exitsMR(N) \la \mr(N,X), \cut(N,X); \\
	&&\notmr(N) \la \naf \exitsMR(N), \cut(N,X); \\
	&&\true(N) \la \emptyT(N), \naf \exitsMR(N); \\
	&&\la \notmr(N), \nemptyT(N); \\
	&&\defeated(N,X) \la \mr(N,X), \mr(N,Y), {\mathrm t}(N,Y), \defeatN(N,Y,X); \\
	&&\la \naf {\mathrm t}(N,X), \naf \defeated(N,X), \mr(N,X)\; \}.
\end{eqnarray*}
With the last module $\pi_\piterate$ we perform Step (iii) of
Proposition~\ref{prop:rbg}. The predicate $\twplus/2$ computes the set
$(T\cap\Pi_F)^\oplus_R$ and with the second rule we start the next
iteration for the framework $(F|_{\Pi_F^C} -(S^+_R \cup
(T\cap\Pi_F)^\oplus_R))$ and the set $(T\cap \Pi_F^C)$. 

\begin{eqnarray*}
 \pi_\piterate &=\{&
	\twplus(N,Y) \la {\mathrm t}(N,X), \mr(N,X), \defeatN(N,X,Y); \\
	&&\argset(M,X) \la \cut(N,X), \naf \mr(N,X), \\
&&\hphantom{\argset(M,X) \la\ } \naf \twplus(N,X), \succ(N,M), \naf \true(N); \\
	&&\inU(M,X) \la {\mathrm t}(N,X), \naf \mr(N,X), \succ(N,M), \naf \true(N)\; \}.
\end{eqnarray*}
\noindent
Finally we put everything together and obtain the program
$\pi_\rground$.

\begin{eqnarray*}
 \pi_\rground &=&
	\pi_\pgroundN \cup \pi_\pcut \cup \pi_\pmr \cup \pi_\pstableN \cup \pi_\piterate.
\end{eqnarray*}
\begin{proposition}
For any AF $F$, $\rground(F)\corr \AS(\pi_\rground(\hat{F}))$.
\end{proposition}

\section{Experimental Evaluation}\label{sec:exp}

In this section we present our results of the performance evaluation. 
We compared the time needed for computing all extensions for the semantics described earlier using both the handcraft saturation-based and the alternative \metasp{} encodings.

The tests were executed on an openSUSE based machine with eight Intel Xeon processors (2.33 GHz) and 49 GB memory. 
For computing the answer sets, we used %
{\tt gringo} (version 3.0.3) for grounding  and the solver {\tt claspD} (version 1.1.1).
The latter being the variant for disjunctive answer-set programs.

\begin{figure}[p!]
	\includegraphics[width=1\textwidth]{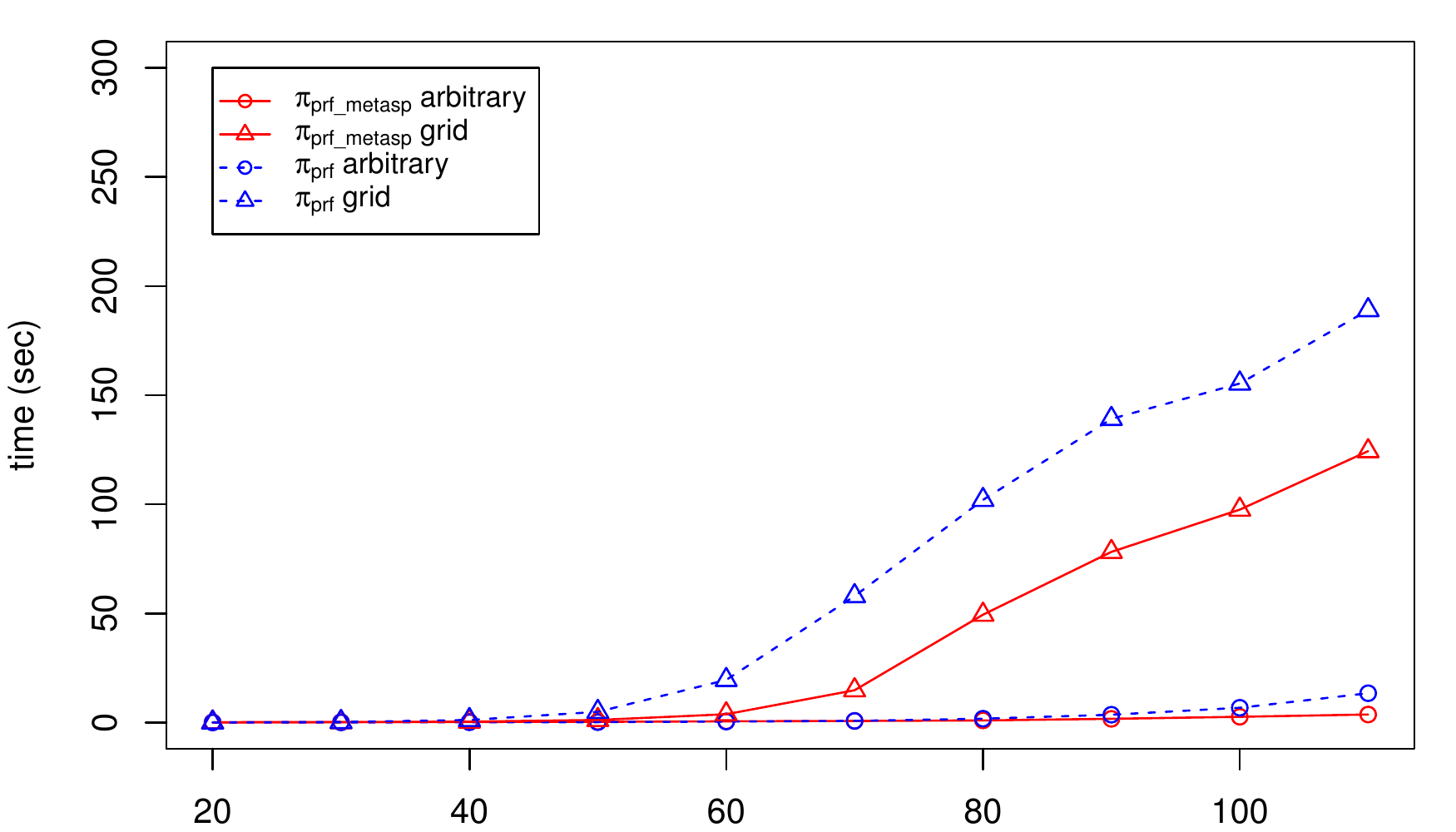}%
	\caption{Average computation time for preferred semantics.}
	\label{fig:p_comp}
\end{figure}
\begin{figure}[p!]
	\includegraphics[width=1\textwidth]{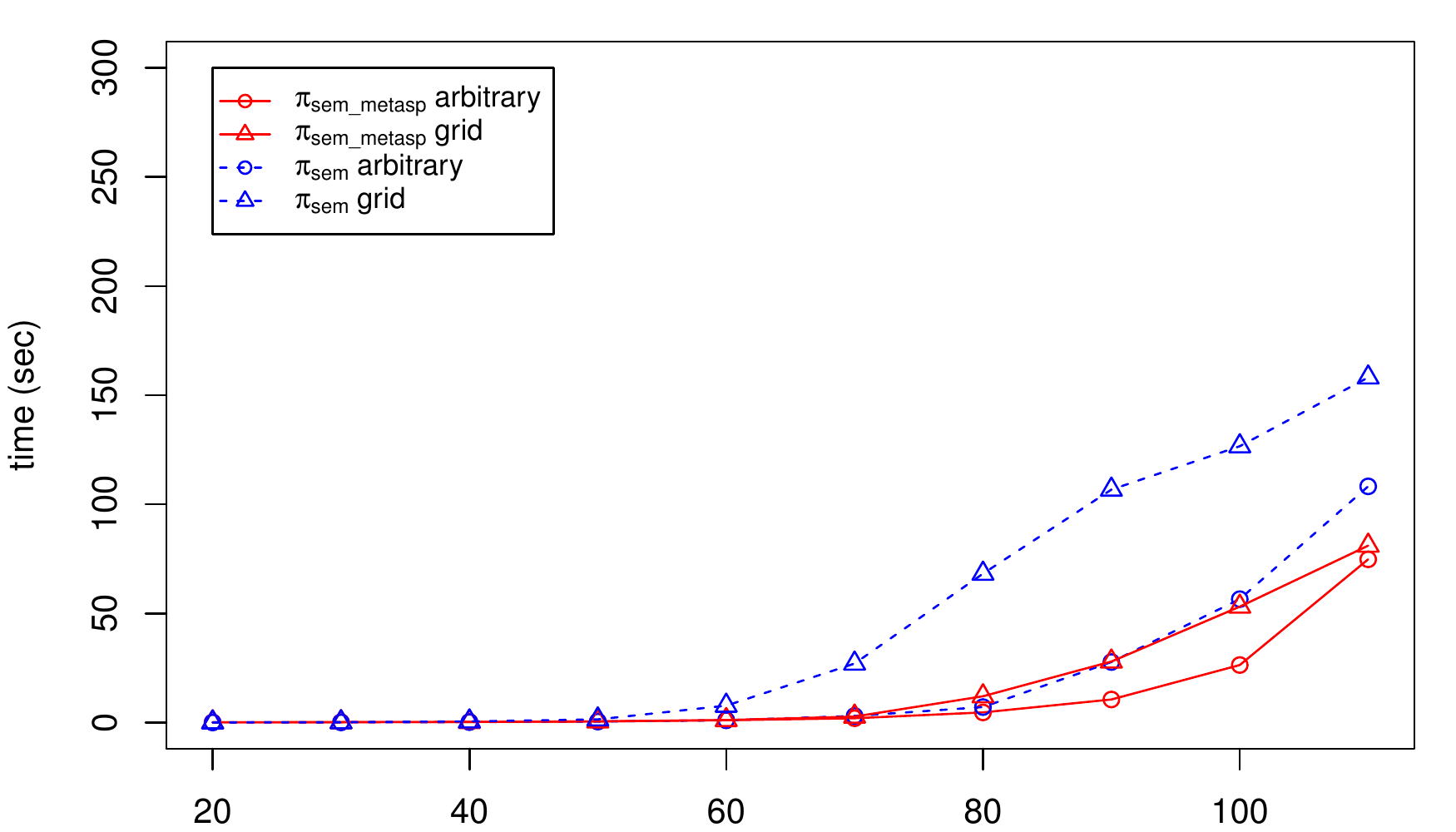}%
	\caption{Average computation time for semi-stable semantics.}
	\label{fig:e_comp}
\end{figure}
\begin{figure}[p!]
	\includegraphics[width=1\textwidth]{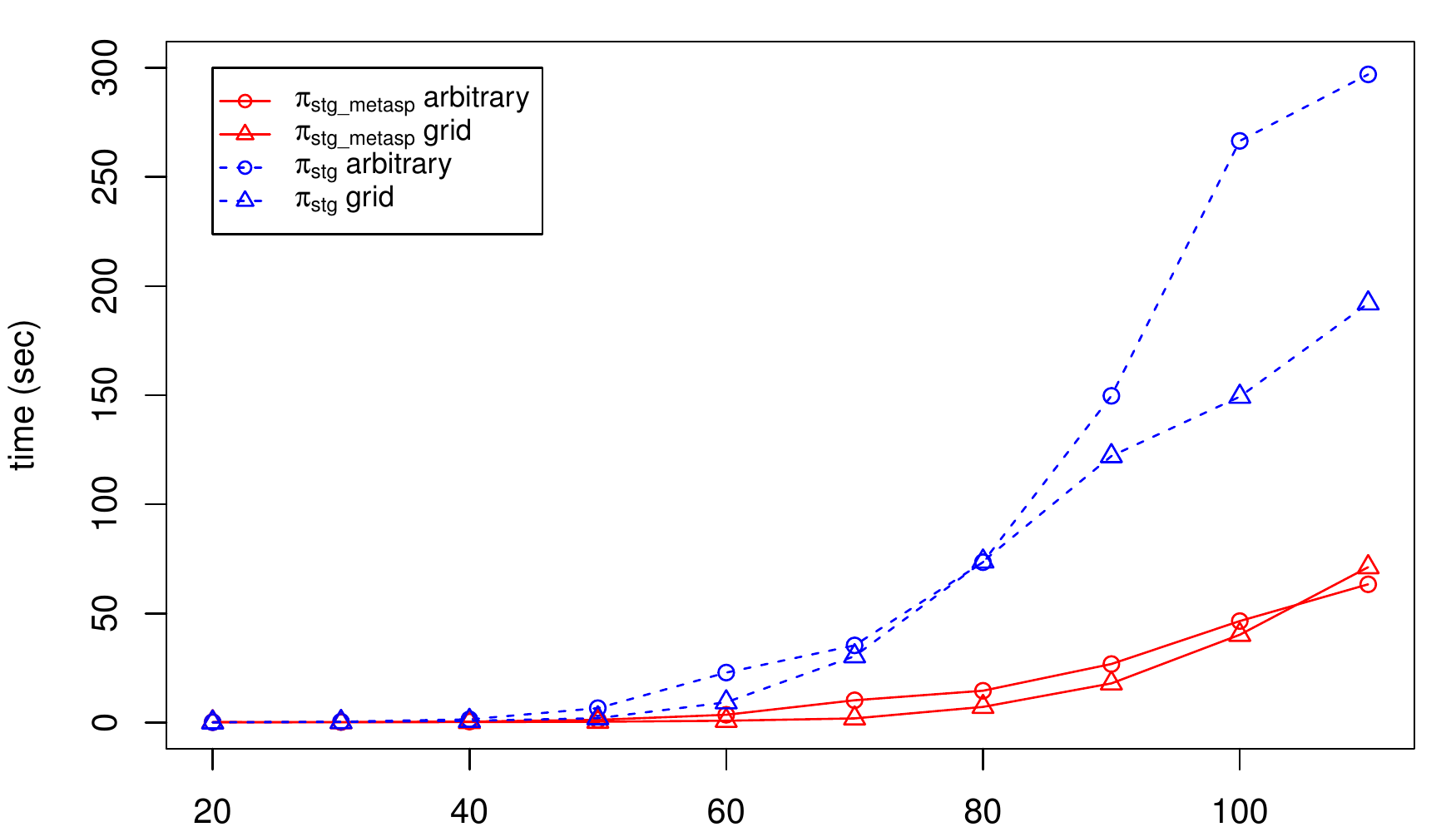}%
	\caption{Average computation time for stage semantics.}
	\label{fig:s_comp}
\end{figure}
\begin{figure}[p!]
	\includegraphics[width=1\textwidth]{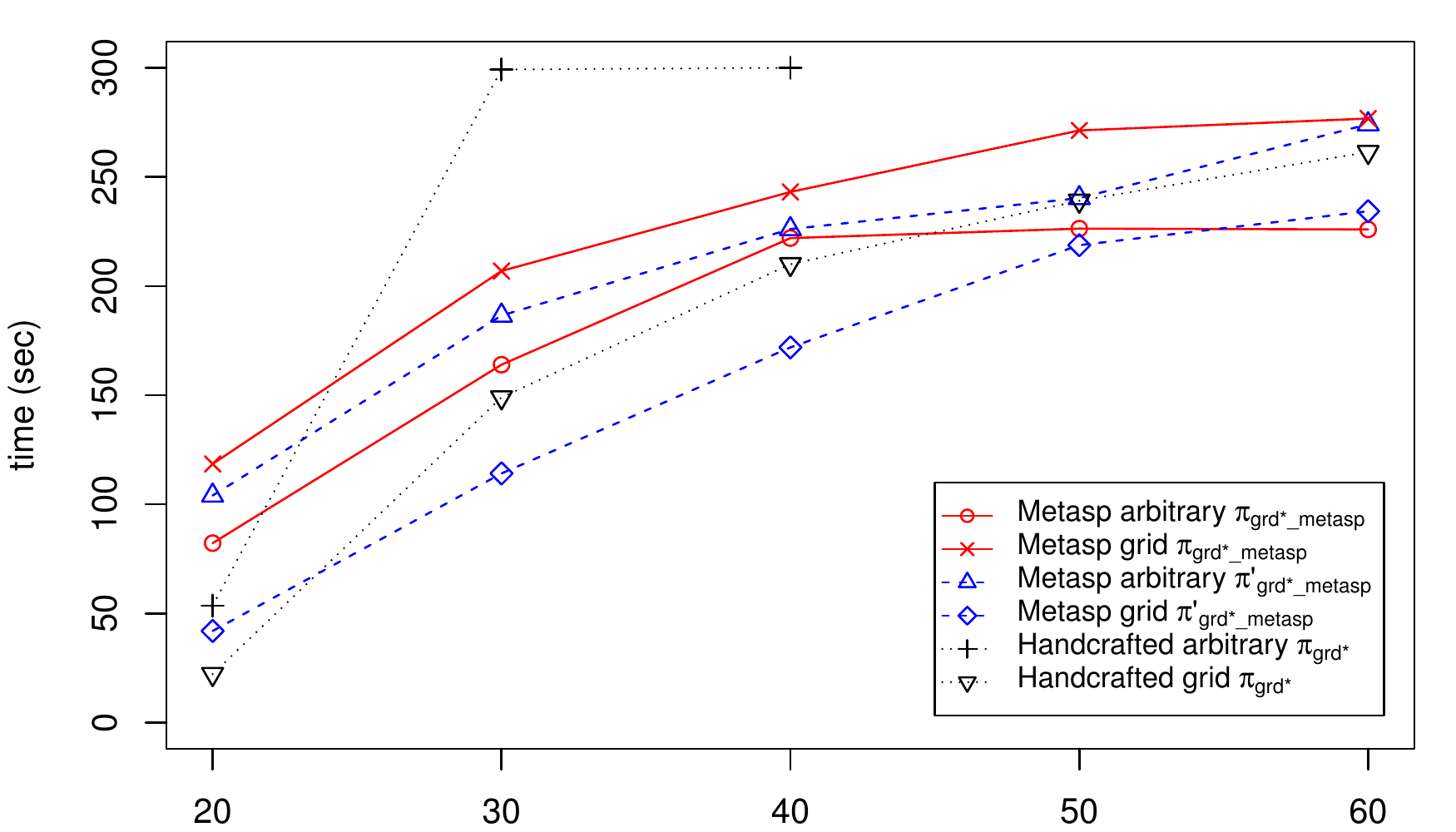}%
	\caption{Average computation time for resolution-based grounded semantics}
	\label{fig:sem_comp2}
\end{figure}

We 
randomly 
generated AFs (i.e.\ graphs) 
ranging from $20$ to $110$ arguments. 
We used two parametrized methods for generating the attack relation.%
The first generates arbitrary AFs and 
inserts for any pair $(a,b)$ the attack from $a$ to $b$ with a given probability $p$. 
The other method generates AFs with a $n \times m$ grid-structure. %
We consider two different neighborhoods, one connecting arguments vertically and horizontally and one that additionally connects the arguments diagonally. Such a connection is a mutual attack with a given probability $p$ and in only one direction otherwise. %
The probability $p$ was chosen between $0.1$ and $0.4$.
Overall 14388 tests were executed, with a timeout of five minutes for each execution. Timed out instances are considered as solved in 300 seconds.
The time consumption was measured using the Linux {\tt time} command. For all the tests we let the solver generate all answer sets, but only outputting the number of models.  To minimize external influences on the test runs, we alternated the different encodings 
during the tests.%

Figures \ref{fig:p_comp} - \ref{fig:s_comp} depict the %
results for the preferred, semi-stable and stage semantics respectively. The figures show the average computation time for both the handcraft and the \metasp{} encoding for a certain number of arguments. We distinguish here between arbitrary, i.e. completely random AFs and grid structured ones.
One can see that the \metasp{} encodings have a better performance, compared to the handcraft encodings.
In particular, for the stage semantics the performance difference between the handcraft and the \metasp{} variant is noticeable. %
Recall that the average computation time includes the timeouts, which strongly influence the diagrams.

For the resolution-based grounded semantics Figure \ref{fig:sem_comp2} shows again the average computation time needed for a certain number of arguments. Let us first consider the case of arbitrary AFs. The handcraft encoding struggled with AFs of size $40$ or larger. Many of those instances could not be solved due to memory faults. This is indicated by the missing data points. Both \metasp{} encodings performed better overall, but still many timeouts were encountered.
If we look more closely at the structured AFs
then we see that $\pi_\rgroundmetasp'$ performs better overall than the other \metasp{} variant. 
Interestingly, computing the grounded part with a handcraft encoding without a Guess\&Check part did not result in a lower computation time on average. The handcraft encoding performed better than $\pi_\rgroundmetasp$ on grids. %

\section{Conclusion}

In this paper, we inspected various ASP encodings 
for four prominent semantics in the area of abstract argumentation.
(1) For the preferred and the semi-stable semantics, we compared 
existing saturation-based encodings \cite{EglyGW10} (here we 
called them handcrafted encodings) with novel alternative encodings
which are based on the recently developed
{\tt metasp}  approach~\cite{GebserKS11}, where subset minimization 
can be directly specified (and a front-end, i.e.\ a meta-interpreter)
compiles such statements back into the core ASP language.
(2) For the stage semantics, we presented here both a
handcrafted and a 
{\tt metasp} encoding. Finally, (3) for the 
resolution-based grounded semantics we provided three 
encodings, two of them using the 
{\tt metasp} techniques. 

Although the {\tt metasp} encodings are much simpler to 
design (since saturation techniques are delegated to the 
meta-interpreter), they perform surprisingly well
when compared with the handcraft encodings which are directly
given to the ASP solver. This shows the practical relevance
of the 
{\tt metasp} technique also in the area of abstract argumentation.
Future work has to focus on further experiments which 
hopefully will
strengthen our observations. 
\end{document}